# Classification of Honey Botanical and Geographical Sources using Mineral Profiles and Machine Learning

Mokhtar Al-Awadhi[1] and Ratnadeep Deshmukh[2]

[1,2] Department of Computer Science and IT, Dr. B.A.M University, Aurangabad, India

[1]mokhtar.awadhi@gmail.com, [2]rrdeshmukh.csit@bamu.ac.in

[1]Corresponding author

**Abstract**

This paper proposes a machine learning-based approach for identifying honey floral and geographical sources using mineral element profiles. The proposed method comprises two steps: preprocessing and classification. The preprocessing phase involves missing-value treatment and data normalization. In the classification phase, we employ various supervised classification models for discriminating between six botanical sources and 13 geographical origins of honey. We test the classifiers' performance on a publicly available honey mineral element dataset. The dataset contains mineral element profiles of honeys from various floral and geographical origins. Results show that mineral element content in honey provides discriminative information useful for classifying honey botanical and geographical sources. Results also show that the Random Forests (RF) classifier obtains the best performance on this dataset, achieving a cross-validation accuracy of 99.30% for classifying honey botanical origins and 98.01% for classifying honey geographical origins.

## 1. Introduction

Honey is a nutrient-dense food with several health advantages. It can be categorized into monofloral and polyfloral based on its botanical origin. Unlike polyfloral honey, monofloral honey derives from a single plant source. Mislabeled honey floral sources frequently deceive consumers. As a result, honey authenticity relies on determining the botanical and geographic origins of the honey to guarantee its quality and safety.

In numerous investigations, various approaches for classifying honey floral sources have been proposed (Al-Awadhi & Deshmukh, 2021, A). Traditionally, beekeepers used melissopalynological analysis to identify honey botanical origins. Using a microscope, pollen grains in honey are examined to determine the honey botanical source. Melissopalynology is often accurate, although it takes a long time and requires much effort. Sample preparation and highly-trained specialists are also required (Corvucci et al., 2015). Modern analytical techniques like hyperspectral imaging spectroscopy with machine learning have been used to classify honey botanical sources (Al-Awadhi & Deshmukh, 2021, B), but this technology does not provide information about trace elements in honey. Honey contains various mineral elements at different concentration. The purpose of this study is to examine the efficacy of Machine Learning (ML) algorithms for categorizing honey floral and geographical origins based on mineral element concentration. This research establishes various ML models, such as Support Vector Machines (SVM), Decision Trees, Linear Discriminant Analysis (LDA), Logistic regression, Quadratic Discriminant Analysis (QDA), and Random Forests. It compares their performance on a public honey mineral element dataset.

Utilizing mineral element profiles for identifying honey floral and geographical sources has been considered in some previous works. For example, (Alina et al., 2021) used Soft-Independent Modeling of Class Analogy (SIMCA) and LDA to classify two geographical origins of honey using isotope ratio analysis coupled with mineral elements. The SIMCA and LDA models achieved classification accuracies of 77% and 85.65% using mineral element profiles. Using isotope coupled with mineral elements, SIMCA and LDA obtained 99% and 98% classification accuracies, respectively. (Squadrone et al., 2020) employed Principal Component Analysis (PCA) and Analysis of Variance (ANOVA) to discriminate between five honey floral sources using 24 mineral elements. The ANOVA analysis showed a significant difference between honey botanical origins using the mineral element content. According to the PCA analysis, the first two PCs explained 77.44% of the variance in the dataset. (Karabagias et al., 2019) utilized LDA and K-Nearest-Neighbors (KNN)-based models for discriminating five botanical origins using mineral element profiles. The authors obtained 67.8% and 78.9% cross-validation accuracies using LDA and KNN, respectively. LDA achieved a cross-validation accuracy of 79.4% using seven mineral elements for classifying various honey botanical origins (Karabagias et al., 2018). Multivariate analysis of variance (MANOVA) and LDA obtained a classification accuracy of 83.8% for classifying four geographical origins of citrus honey using 13 mineral elements (Karabagias et al., 2017).

The dataset utilized in this study consists of 429 instances representing three types of samples: (1) pure honey, (2) adulterated honey, and (3) sugar syrup (Luo, 2020). The falsified honey samples were generated by mixing pure honey samples with sugar syrups at different concentrations. There are six floral sources and 13 geographical origins from which the honey samples come. The floral sources of the honey were Acacia, Rape, Chaste, Linden, Jujube, and T. cochinchinensis (TC). The geographical origins of the pure honey were Liaoning, Shanxi, Jilin, Commodity, Shaanxi, Hebei, Hubei, Henan, Heilongjiang, Jiangxi, Shandong, America, Inner Mongolia. The dataset includes 15 attributes which are 12 mineral element measurements, sample type, botanical origin, region, and level. The mineral elements are Aluminum (Al), Boron (B), Barium (Ba), Zinc (Zn), Calcium (Ca), Strontium (Sr), Iron (Fe), Phosphorus (P), Potassium (K), Sodium (Na), Manganese (Mn), and Magnesium (Mg). Figure 1 and figure 2 show the percentage of the floral and geographical sources of the honey samples in the data set, respectively. The dataset contains missing values for some samples where some mineral elements were not detected (ND). Two previous works on this dataset (Liu, 2021) (Templ & Templ, 2021), but they focused on applying ML techniques to discriminate between pure and impure honey samples. In this study, we use various ML models for classifying the floral and geographical sources of the honey samples.

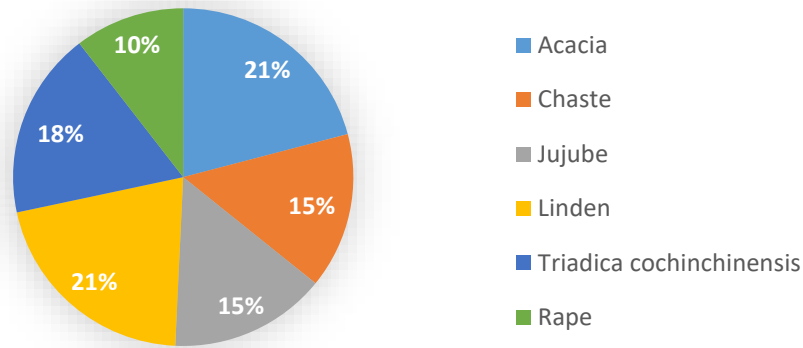

*Figure 1: Percentage of the floral sources of honey samples in the dataset*

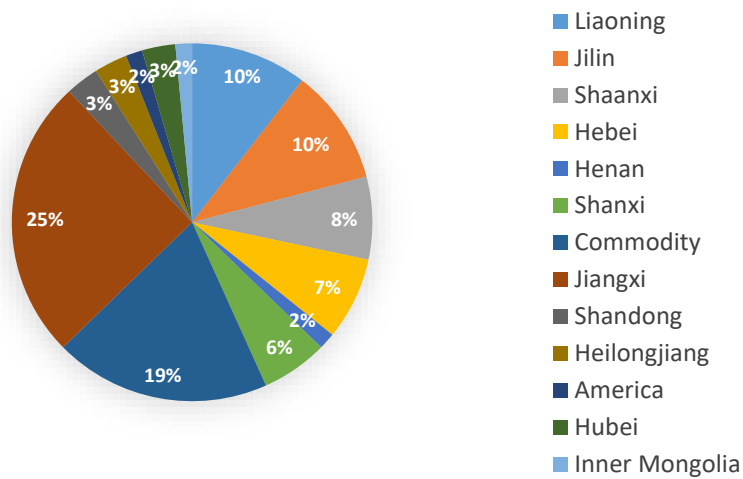

*Figure 2: Percentage of the geographical origins of honey samples in the dataset*

The remainder of this study is structured as follows. Section 2 describes the proposed approach. In Section 3, the experimental results are presented. Section 4 gives an analysis of the acquired results, and Section 5 concludes the paper.

## 2. Proposed Approach

The honey floral source classification method proposed in this paper, depicted in figure 3, comprises two main stages: preprocessing and classification. These two stages are described in detail in the following subsections.

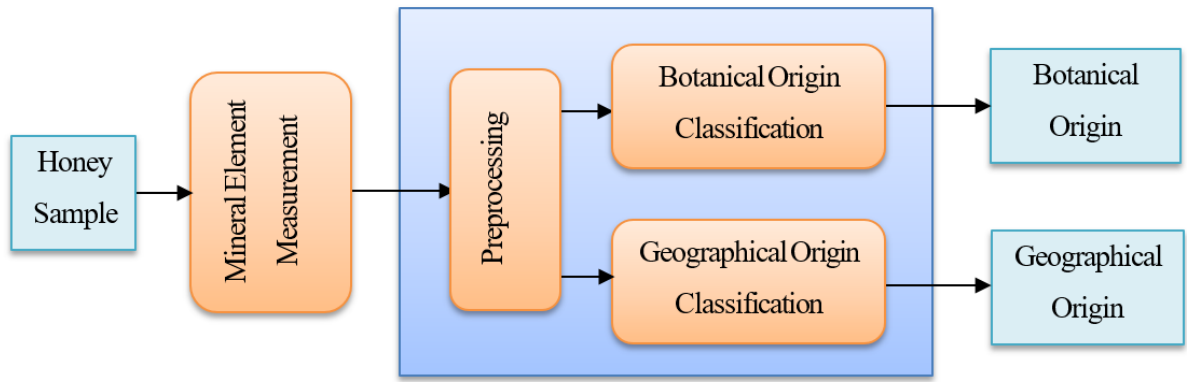

*Figure 3: The suggested system's block diagram*

### 2.1 Preprocessing

This phase consists of two steps. The first step entails filling missing values in the dataset with zeros since these missing values represent the absence of mineral elements in honey samples. Since data normalization helps improve the classification performance of ML models, we normalize the attributes in the dataset to the interval between zero and one using the min-max normalization method (Singh & Singh, 2020) given by equation 1.

$$X'_O = \frac{X_O - \min(X_O)}{\max(X_O) - \min(X_O)} \tag{1}$$

where $X_O$ and $X'_O$ are the original and normalized values, respectively.

### 2.2 Classification

We use various supervised machine learning models to discriminate between honey botanical and geographical origins in this phase. The ML models are SVM, LDA, QDA, Logistic Regression, Decision Tree, and Random Forests. SVM is a supervised ML model first introduced by (Cortes and Vapnik, 1995). It has been successfully used in many classification problems. There are three common functions: linear, polynomial, and radial that can be used as kernel functions of SVM. In addition, SVM is distinguished for its generalization capability and can also be utilized to solve regression problems. LDA is a linear supervised ML model used to extract and classify features. An essential aspect of LDA is changing the original features into a new dimensional feature space where the distance between classes is maximized and the variance with each class is minimized (Tharwat et al., 2017). QDA is a nonlinear variant of LDA suitable for linearly inseparable datasets.

Logistic Regression (LR) is an ML predictive analytic approach that can solve classification tasks based on the probability concept (Tsangaratos & Iliam, 2016). A Decision Tree is a supervised learning model that can solve classification and regression problems. Decision tree-based classifiers are easy to interpret, work with numeric and categorical attributes, and solve binary and multi-class problems (Safavian & Landgrebe, 1991). Like decision trees, Random Forest (RF) is a supervised learning approach capable of carrying out both classification and regression tasks. The RF algorithm is based on the construction of many decision trees. The trees vote for the class with the most votes (Breiman, 2001).

To assess the ML models' performance, we employed 10-fold cross-validation accuracy. Cross-validation has the advantage of avoiding model overfitting. The performance criteria employed were the correct classification accuracy, precision, recall, and F-measure. Precision is the proportion of precisely anticipated positive observations relative to the total number of positive observations expected. We define the recall metric as the proportion of correctly predicted positive observations to all actual class observations. The F-measure represents the weighted mean of precision and recall. Consequently, this Score produces both false positives and false negatives. F-measure is a useful performance metric, especially for uneven data sets.

## 3. Results

We report the achievement of the ML models for identifying honey floral and geographical sources using mineral element profiles in this section.

### 3.1 Classification of Honey Botanical Origins

We assessed how well the ML models performed in distinguishing the botanical origins of honey using the original dataset, which includes instances of pure and impure honey and instances of sugar syrups. The performance of ML models was also evaluated using the pure and impure honey subsets.

#### 3.1.1 Classification of Honey Botanical Origins using Pure Honey Dataset

The classification performance of the models for identifying the botanical origins of honey using the pure honey dataset is shown in Table 1 and Figure 4. Results reveal that the random forest classifier achieves excellent performance for discriminating between honey floral sources, where it got the highest classification accuracy of 99.50% outperforming other classifiers. On the other hand, the SVM classifier achieved the lowest accuracy of 70.65% for classifying the botanical origins of pure honey samples. The logistic regression and decision tree classifiers performed well on this dataset, where they achieved similar classification accuracies near 93%. The linear discriminant analysis classifier did not perform well on this dataset, but the quadratic variant of LDA performed well where it achieved a classification accuracy of 91.54%.

*Table 1: ML classifiers performance for identifying honey floral sources using the pure honey dataset*

| ML Model | Accuracy | Precision | Recall | F-Measure |
|---|---|---|---|---|
| SVM | 70.65 | 0.705 | 0.706 | 0.696 |
| LDA | 75.12 | 0.764 | 0.751 | 0.750 |
| QDA | 91.54 | 0.918 | 0.915 | 0.914 |
| LR | 93.53 | 0.937 | 0.935 | 0.936 |
| DT | 93.03 | 0.935 | 0.930 | 0.931 |
| **RF** | **99.50** | **0.995** | **0.995** | **0.995** |

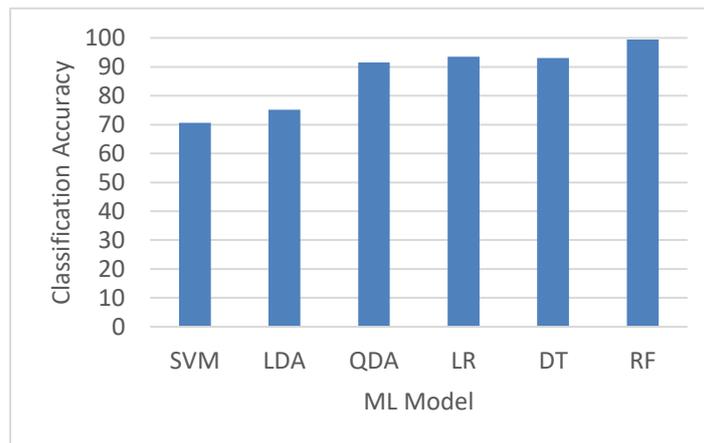

*Figure 4: ML classifiers' performance on the pure honey dataset*

### 3.1.2 Classification of Honey Botanical Origins using Adulterated Honey Dataset.

Table 2 and Figure 5 illustrate the achievement of the machine learning algorithms for identifying honey floral sources using the adulterated honey sample dataset. The random forests classifier was superior to its performance using the pure honey dataset, achieving the highest classification accuracy of 99.44%. The performance of the SVM model has improved on this data set, where it obtained an accuracy of 99.22%, outperforming the decision tree classifier. The LDA classifier obtained an accuracy of 92.22%, similar to the SVM classifier, but the quadratic variant of LDA got a better accuracy of 94.44%. The logistic regression classifier obtained an outstanding performance of 97.22% on this data subset, outperforming its performance on the pure honey sample dataset.

*Table 2 ML classifiers performance for identifying honey floral sources using the impure honey dataset*

| ML Model | Accuracy | Precision | Recall | F-Measure |
|---|---|---|---|---|
| SVM | 92.22 | 0.928 | 0.922 | 0.922 |
| LDA | 92.22 | 0.923 | 0.922 | 0.923 |
| QDA | 94.44 | 0.951 | 0.944 | 0.946 |
| LR | 97.22 | 0.973 | 0.972 | 0.972 |
| DT | 87.22 | 0.876 | 0.872 | 0.873 |
| **RF** | **99.44** | **0.995** | **0.994** | **0.994** |

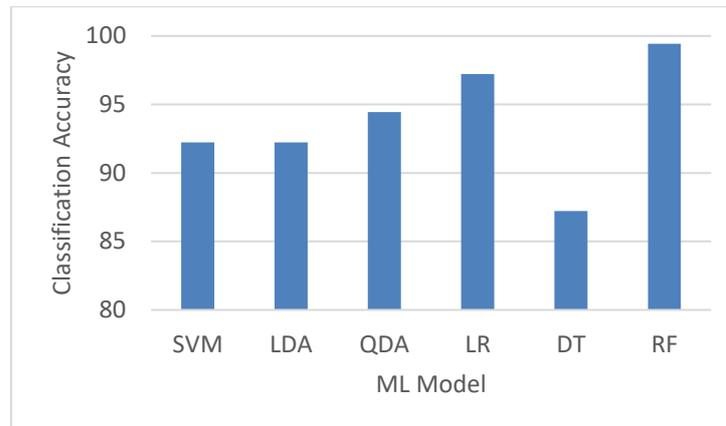

*Figure 5: ML classifiers' performance on the adulterated honey dataset*

### 3.1.3   Classification Honey Botanical Origins using Original Dataset

Table 3 and figure 6 show the achievement of the ML classifiers for identifying honey floral sources using the original dataset, which contains instances of pure and impure honey samples and instances of sugar syrup. Generally, the ML models show outstanding performance in identifying honey floral sources using mineral element profiles. The RF-based classifier achieved the highest prediction accuracy of 99.30%, outperforming other classifiers. The accuracy achieved by the random forest classifier on this dataset is slightly lower than the accuracy achieved on the pure and impure honey sample datasets since the original dataset includes instances of sugar syrup samples and pure and impure honey samples. The LDA classifier obtained the lowest accuracy of 75.82% on this dataset, but the nonlinear LDA obtained better performance indicating the dataset is linearly inseparable. The decision tree classifier obtained a cross-validation accuracy of 94.84%, similar to the accuracy obtained by QDA.

*Table 3: ML classifiers performance for identifying honey floral sources using the original dataset*

| ML Model | Accuracy | Precision | Recall | F-Measure |
|---|---|---|---|---|
| SVM | 77.93 | 0.793 | 0.779 | 0.779 |
| LDA | 75.82 | 0.779 | 0.758 | 0.762 |
| QDA | 94.60 | 0.947 | 0.946 | 0.946 |
| LR | 90.85 | 0.909 | 0.908 | 0.909 |
| DT | 94.84 | 0.949 | 0.948 | 0.940 |
| **RF** | **99.30** | **0.993** | **0.993** | **0.993** |

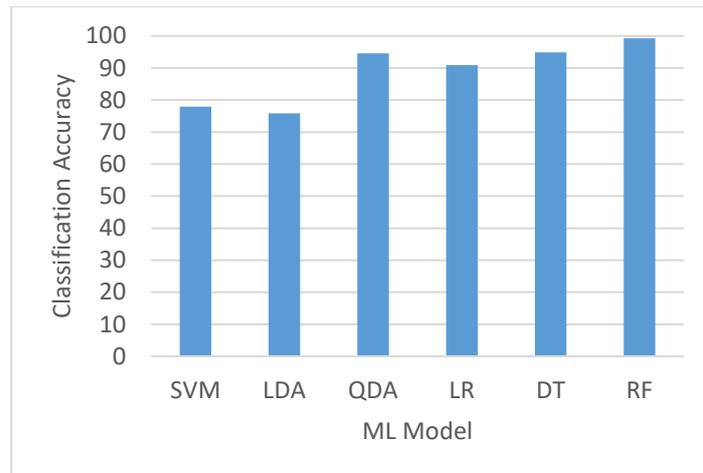

*Figure 6: ML classifiers' performance on the original dataset*

Table 4 displays the detailed performance of the RF model for each floral source. The results demonstrate that the RF-based classifier successfully classifies the acacia, chaste, and TC honey samples and sugar syrup samples with 100% accuracy. Besides that, the random forest classifies honey samples from other botanical origins with a high accuracy near 98%.

*Table 4: Random forests detailed performance by class*

| Class | TP Rate | FP Rate | Precision | Recall | F-Measure |
|---|---|---|---|---|---|
| Acacia | 1.000 | 0.000 | 1.000 | 1.000 | 1.000 |
| Chaste | 1.000 | 0.003 | 0.984 | 1.000 | 0.992 |
| Jujube | 0.983 | 0.000 | 1.000 | 0.983 | 0.992 |
| Linden | 0.986 | 0.000 | 1.000 | 0.986 | 0.993 |
| Rape | 0.980 | 0.003 | 0.980 | 0.980 | 0.980 |
| TC | 1.000 | 0.003 | 0.985 | 1.000 | 0.992 |
| Syrup | 1.000 | 0.000 | 1.000 | 1.000 | 1.000 |

## 3.2 Classification of Honey Geographical Origins

In this subsection, we present the achievement of the ML models for discriminating between the geographical origins of honey. We used the pure honey sample dataset to evaluate the effectiveness of the ML algorithms since the geographical regions of the adulterated honey samples were not provided in the original dataset. Table 5 and figure 7 show the achievement of the ML classifiers for discriminating between the geographical sources of honey. The results indicate that the random forests model achieved the best performance with a cross-validation accuracy of 98.01% for discriminating between honey geographical origins, outperforming other classifiers on this dataset. While SVM and the two variants of LDA performed poorly on this dataset, the decision tree classifier performed better with a correct classification accuracy of 83.58%. Some precision and f-sore values are missing for the SVM and QDA since these two classifiers could not classify instances of honey samples from some geographical origins.

*Table 5: Performance of ML models for classifying honey samples to their geographical origins*

| ML Model | Accuracy | Precision | Recall | F-Measure |
|---|---|---|---|---|
| SVM | 42.79 | ? | 0.428 | ? |
| LDA | 57.71 | 0.611 | 0.577 | 0.581 |
| QDA | 62.69 | ? | 0.627 | ? |
| LR | 76.62 | 0.781 | 0.766 | 0.765 |
| DT | 83.58 | 0.836 | 0.836 | 0.833 |
| **RF** | **98.01** | **0.981** | **0.980** | **0.980** |

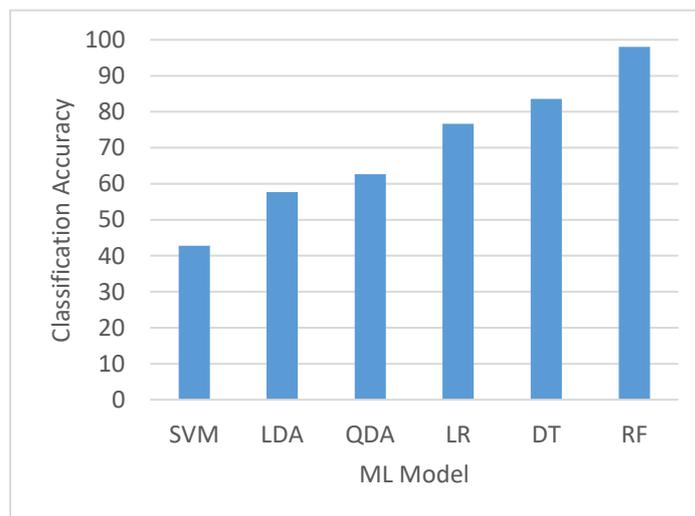

*Figure 7: Performance of ML models for classifying honey samples to their geographical origins*

Table 6 displays the effectiveness of the Random Forests model for identifying the geographical sources of honey samples in the dataset. The results demonstrate that the random forests model accurately classifies nine geographical origins with an accuracy of 100%. Besides that, the random forests model classifies three geographical origins with high accuracy near 97%. The random forests model classifies the American geographical origin with a low accuracy of 67%.

*Table 6: Detailed accuracy of RF classifier for each geographical origin*

| Class | TP Rate | FP Rate | Precision | Recall | F-Measure |
|---|---|---|---|---|---|
| **Liaoning** | 1.000 | 0.011 | 0.913 | 1.000 | 0.955 |
| **Jilin** | 0.952 | 0.000 | 1.000 | 0.952 | 0.976 |
| **Shaanxi** | 1.000 | 0.000 | 1.000 | 1.000 | 1.000 |
| **Hebei** | 0.933 | 0.000 | 1.000 | 0.933 | 0.966 |
| **Henan** | 1.000 | 0.000 | 1.000 | 1.000 | 1.000 |
| **Shanxi** | 1.000 | 0.000 | 1.000 | 1.000 | 1.000 |
| **Commodity** | 0.974 | 0.006 | 0.974 | 0.974 | 0.974 |
| **Jiangxi** | 1.000 | 0.007 | 0.981 | 1.000 | 0.990 |
| **Shandong** | 1.000 | 0.000 | 1.000 | 1.000 | 1.000 |
| **Heilongjiang** | 1.000 | 0.000 | 1.000 | 1.000 | 1.000 |
| **America** | 0.667 | 0.000 | 1.000 | 0.667 | 0.800 |
| **Hubei** | 1.000 | 0.000 | 1.000 | 1.000 | 1.000 |
| **Inner Mongolia** | 1.000 | 0.000 | 1.000 | 1.000 | 1.000 |

## 4. Discussion

This research employed several supervised ML algorithms for classifying honey floral and geographical sources using mineral element profiles. Before classification, we filled missing values of some attributes with zeros since these missing values represent the absence of mineral elements in some honey and sugar syrup samples. After that, we normalized the dataset attributes since normalization speeds up ML model training and helped produce consistent results. To classify honey floral and geographical origins, we used linear classifiers, such as linear-kernel SVM, LDA, and logistic regression, and nonlinear classification models, such as decision trees and random forests. Generally, results show that ML models show an outstanding performance in classifying honey botanical and geographical origins using mineral element profiles. The nonlinear classifiers outperformed linear classifiers confirming that the dataset is linearly inseparable. The random forests model had the best cross-validation accuracy when distinguishing between botanical and geographic sources. Most classifiers, except random forests, performed well for identifying honey botanical sources but performed poorly in recognizing honey geographical origins because floral sources in the dataset were fewer than the geographical origins. Results show that ML models can identify honey botanical origins whether the honey samples were pure or impure with various sugar syrups. Table 7 compares the results of the proposed approach to the results obtained in previous work. Results reveal that the method proposed in this study achieves the highest cross-validation accuracy for classifying honey botanical and geographical origins, outperforming the models developed in previous work.

*Table 7: Comparison with previous work for classifying honey botanical and geographical origins*

[1]*BO: Botanical Origin,* [2]*GO: Geographical Origin,* [3]*ME: Mineral Elements,* [4]*GOC: Geographical Origin Classification,* [5]*BOC: Botanical Origin Classification.*

| Reference | Aim | BO | GO | ME | ML Models | Accuracy |
|---|---|---|---|---|---|---|
| D. Alina et al. [4] | GOC[4] | 12 | 2 | 7 | SIMCA | 77% |
|  |  |  |  |  | LDA | 85.65% |
| S. Squadrone et al. [5] | BOC[5] | 5 | - | 24 | PCA | VAR (PC1, PC2) = 77.44 % |
|  |  |  |  |  | ANOVA |  |
| I. K. Karabagias et al. [6-8] | BOC | 5 | 5 | 9 | LDA | 67.8% |
|  |  |  |  |  | KNN | 78.9% |
|  | BOC | various | - | 7 | LDA | 79.4% |
|  | GOC | - | 4 | 13 | LDA | 83.8% |
| **Proposed System** | **BOC** | 6 | 13 | 11 | **Random Forests** | BO = 99.30% |
|  | **GOC** |  |  |  |  | GO = 98.01% |

## 5. Conclusion

This research established a machine learning-based approach for distinguishing honey floral and geographical origins using mineral element profiles. The developed system consists of two stages: preprocessing and classification. The preprocessing stage involves filling missing values and normalization. We used several supervised ML models for discriminating between honey botanical and geographical origins in the classification phase. Experimental findings show that mineral element profiles provide robust discriminative information that machine learning models can classify various honey botanical and geographical origins. Results also reveal that adulterating honey by different sugar syrups has no significant impact on the performance of ML models for classifying honey flora and geographical sources. Findings also demonstrate that the random forests classifier achieves the highest classification accuracy on this dataset for discriminating between honey botanical and geographical origins.

## Acknowledgments

The Department of Science and Technology (DST) funded this research through the Funds for Infrastructure under Science and Technology (FIST) with grant number SR/FST/ETI-340/2013 to the Department of Computer Science and Information Technology at Dr. Babasaheb Ambedkar Marathwada University in Aurangabad, Maharashtra, India. The authors would like to thank the department and university officials for providing the resources and assistance needed to perform the study.